\documentclass{article}
\PassOptionsToPackage{numbers, compress}{natbib}
\usepackage[final]{nips_2018}
\usepackage[utf8]{inputenc}
\usepackage{amsfonts}
\usepackage{nicefrac}
\usepackage{amsmath}
\usepackage{amssymb}
\usepackage{amsthm}
\usepackage{amstext}
\usepackage{cleveref}
\usepackage{acronym}
\usepackage{bm}
\usepackage{url}
\usepackage{graphicx}
\usepackage{subcaption}
\usepackage[inline]{enumitem}
\usepackage{wrapfig}
\usepackage{pdfpages}
\usepackage{lipsum}
\usepackage{appendix}

\title{Facilitating Bayesian Continual Learning by Natural Gradients and Stein Gradients}

\author{
  Yu Chen\thanks{This work  was conducted during Yu Chen's internship at Amazon.} \\
  University of Bristol\\
  \texttt{yc14600@bristol.ac.uk} \\
   \And
   Tom Diethe \\
   Amazon \\
   \texttt{tdiethe@amazon.com} \\
   \And
   Neil Lawrence \\
   Amazon \\
   \texttt{lawrennd@amazon.com} \\
}

\begin{document}

\maketitle

\acrodef{KL}{Kullback-Leibler}
\acrodef{VI}{Variational Inference}
\acrodef{ELBO}{Evidence Lower Bound}
\acrodef{MAML}{Model-Agnostic Meta-Learning for Fast Adaptation of Deep Networks}
\acrodef{VAE}{Variational Auto-Encoder}
\acrodef{IAF}{Inverse Autoregressive Flow}
\acrodef{MAML-HB}{\ac{MAML}-Hierarchical Bayes}
\acrodef{K-FAC}{Kronecker-factored approximate curvature}
\acrodef{BMAML}{Bayesian \ac{MAML}}
\acrodef{SVGD}{Stein Variational Gradient Descent}
\acrodef{EWC}{Elastic Weight Consolidation}
\acrodef{SI}{Synaptic Intelligence}
\acrodef{VCL}{Variational Continual Learning}
\acrodef{IWAE}{Importance Weighted Auto Encoder}
\acrodef{NF}{Normalizing Flow}
\acrodef{Real NVP}{Real-valued Non-Volume Preserving}
\acrodef{NN}{Neural Network}
\acrodef{BNN}{Bayesian \ac{NN}}
\acrodef{SGD}{Stochastic Gradient Descent}
\acrodef{GNG}{Gaussian Natural Gradient}
\acrodef{MSE}{Mean Squared Error}
\acrodef{RKHS}{Reproducing Kernel Hilbert Space}
\acrodef{RBF}{Radial Basis Function}

\begin{abstract}
Continual learning aims to enable machine learning models to learn a general solution space for past and future tasks in a sequential manner. Conventional models tend to forget the knowledge of previous tasks while learning a new task, a phenomenon known as \textit{catastrophic forgetting}. When using Bayesian models in continual learning, knowledge from previous tasks can be retained in two ways:
\begin{enumerate*}[label=(\roman*)]
    \item posterior distributions over the parameters, containing the knowledge gained from inference in previous tasks, which then serve as the priors for the following task;
    \item \emph{coresets}, containing knowledge of data distributions of previous tasks. 
\end{enumerate*}
Here, we show that Bayesian continual learning can be facilitated in terms of these two means through the use of natural gradients and Stein gradients respectively. 

\end{abstract}

\section{Background}
There are several existing approaches for preventing \textit{catastrophic forgetting} of regular (non-Bayesian) \acp{NN} by constructing a regularization term from parameters of previous tasks, such as \ac{EWC} \cite{kirkpatrick2017overcoming} and \ac{SI} \cite{zenke2017continual}. In the Bayesian setting, \ac{VCL} \cite{nguyen2017variational} proposes a framework that makes use of \ac{VI}.
\begin{equation}\label{eq:vcl_obj}
        \mathcal { L } _ { \mathrm { VCL } } \left(  { \theta }  \right) =  \mathbb { E } _ { q _ { t } ( \mathbf { \theta } ) } \left[ \log p \left( \mathcal{D} _ { t } | \mathbf { \theta } \right) \right] - \mathrm { KL } \left( q _ { t } ( \mathbf { \theta } ) \| q _ { t - 1 } ( \mathbf { \theta } ) \right).
\end{equation}
The objective function is as in \Cref{eq:vcl_obj}, where $t$ is the index of tasks, $q_t(\theta)$ represents the approximated posterior of parameters $\theta$ of task $t$ and $D_t$ is the data of task $t$. This is the same as the objective of conventional \ac{VI} except the prior is from the previous task, which produces the regularization by \ac{KL}-divergence between parameters of current and previous tasks. In addition, \ac{VCL} \cite{nguyen2017variational} proposes a predictive model trained by coresets of seen tasks to performs prediction for those tasks, where the coresets consist of data samples from the dataset of each seen task except the training data $D_t$ of each task.
\begin{equation}\label{eq:vcl_cs}
    \begin{split}
        \hat{q}_t = \arg\max_{q_t} \mathbb { E } _ { q _ { t } ( \mathbf { \theta } ) } \left[ \log p \left( \mathcal{C} _ { t } | \mathbf { \theta } \right) \right] - \mathrm { KL } \left( q _ { t } ( \mathbf { \theta } ) \| q^* _ { t } ( \mathbf { \theta } ) \right).
    \end{split}
\end{equation}
As shown in \Cref{eq:vcl_cs}, $C_t = \{c_1,c_2,\dots,c_t\}$ represents the collection of coresets at task $t$ and $q_t^*(\theta)$ is the optimal posterior obtained by \Cref{eq:vcl_obj}. \ac{VCL} shows promising performance comparing to \ac{EWC} \cite{kirkpatrick2017overcoming} and \ac{SI} \cite{zenke2017continual}, which demonstrates that effectiveness of Bayesian approaches to continual learning.

\section{Facilitating Bayesian continual learning by natural gradients}
In order to prevent \textit{catastrophic forgetting} in Bayesian continual learning, we would prefer the posteriors of a new task stay as close as possible to the posteriors of the previous task. Conventional gradient methods give the direction of steepest descent of parameters in Euclidean space, which might cause a large difference in terms of distributions following a small change in terms of parameters. We posit that natural gradient methods may be a better choice than the conventional gradient descent. The definition of the natural gradient is the direction of steepest descent in Riemannian space rather than Euclidean space, which means the natural gradient would prefer the smallest change in terms of distribution while optimizing some objective function  \cite{pascanu2013revisiting}. The formulation is as below:
\begin{equation}\label{eq:ngrad}
    \begin{split}
         \hat{\nabla} \mathcal { L } ( \beta ) & = \nabla \mathcal { L } ( \beta ) \mathbb { E } _ {  { \theta } } \left[ \left( \nabla \log p (  { \theta } | \beta ) \right) ^ { T } \left( \nabla \log p (  { \theta | \beta} ) \right) \right] ^ { - 1 }   
          \stackrel { \text { def } } { = } \nabla \mathcal { L } ( \beta ) \mathbf { F }_{\beta} ^ { - 1 } ,
    \end{split}
\end{equation}
where $\mathbf{F}_{\beta}$ is the Fisher information of $\beta$. 

\subsection{Natural gradients of the exponential family}

Specifically, when the posterior of a parameter $\theta$ is from the exponential family, we can write it in the form of $p(\theta|\beta) = h(\theta)\exp\{\eta(\beta)^T u(\theta) - a(\eta(\beta))\}$, 
where $h(\cdot)$ is the base measure and $a(\cdot)$ is log-normalizer, $\eta(\cdot)$ is natural parameter and $u(\cdot)$ are sufficient statistics. Then the Fisher information of $\beta$ is the covariance of the sufficient statistics which is the second derivative of $a(\cdot)$ \cite{hoffman2013stochastic}:
\begin{equation}
    \begin{split}
        \mathbf{F}_{\beta} = \mathbb{E}_{q}[(u(\theta) - \mathbb{E}_q[u(\theta)])^T(u(\theta)-\mathbb{E}_q[u(\theta)])]
        = \nabla_{\beta}^2 a(\eta(\beta)).
    \end{split}
\end{equation}
In this case, the natural gradient of $\beta$ is the transformation of the Euclidean gradient by the precision matrix of the sufficient statistics $u(\theta)$.

\subsection{Gaussian natural gradients and the Adam optimizer}\label{sec:gng}

In the simplest (and most common) formulation of \acp{BNN}, the weights are drawn from Gaussian distributions, with a mean-field factorization which assumes that the weights are independent. Hence, we have an approximate posterior for each weight $q(\theta_i|\mu_i,\sigma_i) = \mathcal{N}(\mu_i,\sigma_i^2)$,
where $\mu_i, \sigma_i$ are the parameters to be optimized,
and their Fisher information has an analytic form:
\begin{equation}\label{eq:gfisher}
    \begin{split}
        & \mathbf{F}_{\mu_i} = 
        {1}/{\sigma_i^2}, \quad
         \mathbf{F}_{\mathbf{v}_i} = 2 
         , \quad \text{where} \ \ \mathbf{v}_i = \log \sigma_i .
    \end{split}
\end{equation}
%
Consequently, the natural gradient of the mean of posterior can be computed as follows, where $\mathcal{L}$ represents the objective (loss) function:
\begin{equation}\label{eq:mu_ng}
    \begin{split}
        \hat{g}_{\mu_i}  = \sigma_i^2 g_{\mu_i}, \quad g_{\mu_i} = \nabla_{\mu_i}\mathcal{L}.
    \end{split}
\end{equation}

\Cref{eq:mu_ng} indicates that small $\sigma_i$ can cause the magnitude of natural gradient to be much reduced. In addition, \acp{BNN} usually need very small variances in initialization to obtain a good performance at prediction time, which brings difficulties when tuning learning rates when applying vanilla \ac{SGD} to this \acf{GNG}.
As shown in Figure 1 and 2 in the supplementary materials we can see how the scale of variance in initialization changes the magnitude of \ac{GNG}. 

Meanwhile, Adam optimization \cite{kingma2014adam} provides a method to ignore the scale of gradients in updating steps, which could compensate this drawback of \ac{GNG}. 
More precisely, Adam optimization uses the second moment of gradients to reduce the variance of updating steps:
\begin{equation}\label{eq:adam}
    \begin{split}
        \theta_{k} & \leftarrow \theta_{k-1} - \alpha_k * {\mathbb{E}_k[\mathbf{g}_{\theta,k}]}\bigm/{\sqrt{\mathbb{E}_k[\mathbf{g}_{\theta,k}^T \mathbf{g}_{\theta,k}}]},
        \quad
        \mathbf{g}_{\theta} = \nabla_{\theta} \mathcal{L},
    \end{split}
\end{equation}
where $k$ is index of updating steps, $\mathbb{E}_k$ means averaging over updating steps, $\alpha_k$ is the adaptive learning rate at step $k$. 
Considering the first and second moments of $\hat{g}_{\mu_i,k}$, 
\begin{equation}\label{eq:ng_moments}
    \begin{split}
         \mathbb{E}_k[\hat{g}_{\mu_i,k}] 
        & = 
        \mathbb{E}_k[\sigma_{i,k}^2] \mathbb{E}_k[g_{\mu_i,k}] + \text{cov}(\sigma^2_{i,k},g_{\mu_i,k})\\
         \mathbb{E}_k[\hat{g}_{\mu_i,k}^2]
        &= (\mathbb{E}_k[\sigma^2_{i,k}]^2 + \text{var}(\sigma^2_{i,k}))\mathbb{E}_k[g_{\mu_i,k}^2] + \text{cov}(\sigma^4_{i,k},g_{\mu_i,k}^2)
    \end{split}
\end{equation}
We can see that only when $\text{var}(\sigma_{i,k}^2) = 0$ and $g_{\mu_i,k}$ are independent from $\sigma_{i,k}^2$, the updates of \ac{GNG} are equal to the updates by Euclidean gradients in the Adam optimizer. It also that shows larger $\text{var}(\sigma_{i,k}^2)$ will result in smaller updates when applying Adam optimization to \ac{GNG}.

We show comparison between different gradient descent algorithms in the supplementary materials. More experimental results are shown in  \Cref{sec:experiments}. 

In \textbf{non-Bayesian} models, natural gradients may have problems with the Adam optimizer because there is no posterior $p(\theta|\beta)$ defined. The distribution measured in natural gradient is the conditional probability $p(x|\theta)$ \cite{pascanu2013revisiting} and the loss function is usually $\mathcal{L}_{\theta} = \mathbb{E}_x[\log p(x|\theta)]$. In this case the natural gradient of $\theta$ becomes:
\begin{equation}\label{eq:nbys_ng}
    \begin{split}
        \hat{\mathbf{g}}_{\theta} = \frac{\mathbb{E}_{x}[g_\theta]}{\mathbb{E}_{x}[g_\theta^T g_\theta]},
        \quad
        {g}_{\theta} = \nabla_{\theta} \log p(x|\theta),
        \quad
        \nabla_{\theta} \mathcal{L} = \mathbb{E}_x[g_\theta].
    \end{split}
\end{equation}
If we apply this to the Adam optimizer, which means replacing $\mathbf{g}_{\theta}$  in \Cref{eq:adam} by $\hat{\mathbf{g}}_{\theta}$,
the formulation is duplicated and involves the fourth moment of the gradient, which is undesirable for both Adam optimization and natural gradients. One example is \ac{EWC} \cite{kirkpatrick2017overcoming} which uses Fisher information to construct the penalty of changing previous parameters, hence, it has a similar form with Adam and it works worse with Adam than with vanilla \ac{SGD} in our experience.
However, this is not the case for \textbf{Bayesian} models, where \Cref{eq:nbys_ng} does not hold because the parameter $\theta$ has its posterior $p(\theta|\beta)$ and then the loss function $\mathcal{L}$ is optimized w.r.t. $\beta$, meanwhile $\nabla_{\beta} \mathcal{L} \neq \mathbb{E}_{\theta}[\nabla_\beta \log p(\theta | \beta)]$ in common cases. 

\section{Facilitating Bayesian Continual Learning with Stein Gradients}

In the context of continual learning, ``coresets'' are small collections of data samples of every learned task, used for task revisiting when learning a new task \cite{nguyen2017variational}. The motivation is to retain summarized information of the data distribution of learned tasks so that we can use this information to construct an optimization objective for preventing parameters from drifting too far away from the solution space of old tasks while learning a new task. Typically, the memory cost of coresets will increase with the number of tasks, hence, we would prefer the size of a coreset as small as possible.
There are some existing approaches to Bayesian coreset construction for scalable machine learning \cite{huggins2016coresets,campbell2018bayesian}, the idea is to find a sparse weighted subset of data to approximate the likelihood over the whole dataset. In their problem setting the coreset construction is also crucial in posterior approximation, and the computational cost is at least $O(MN)$ \cite{campbell2018bayesian}, where $M$ is the coreset size and $N$ is the dataset size. In Bayesian continual learning, the coreset construction does not play a role in the posterior approximation of a task. For example, we can construct coresets without knowing the posterior, i.e. random coresets, $K$-centre coresets \cite{nguyen2017variational}. However, the information of a learned task is not only in its data samples but also in its trained parameters,
so we consider constructing coresets using our approximated posteriors, yet without intervening the usual Bayesian inference procedure.

\subsection{Stein gradients}
Stein gradients \cite{liu2016stein} can be used to generate samples of a known distribution. Suppose we have a series of samples $\mathbf{x}^{(l)}$ from the empirical distribution $p(\mathbf{x})$, and we update them iteratively to move closer to samples from the conditional distribution $p(\mathbf{x}|\theta)$ by $\mathbf{x}_{k+1}^{(l)} = \mathbf{x}_{k}^{(l)} + \bm{\epsilon} \phi^*(\mathbf{x}_k^{(l)})$, where
\begin{equation}
    \begin{split}
        & \phi^* = \arg \max_{\phi \in \mathcal{F}} - \nabla_{\bm{\epsilon}} D_{KL}(p(\mathbf{x})||p(\mathbf{x|\theta}))|_{\bm{\epsilon}=0} = \arg \max_{\phi \in \mathcal{F}} \mathbb{E}_q[\mathcal{A}_p \phi(\mathbf{x})].
    \end{split}
\end{equation}  
$\bm{\phi}^*(\cdot)$ is chosen to decrease the \ac{KL}-divergence between $p(\mathbf{x})$ and $p(\mathbf{x}|\theta)$ in the steepest direction; $\mathcal{F}$ is chosen to be the unit ball of the \acl{RKHS} to give a closed form update of samples; and $\mathcal{A}_p$ is Stein operator.
Thus, the Stein gradient can be computed by:
\begin{equation}\label{eq:stein_grads2}
    \begin{split}
        \phi^*(\mathbf{x}_k^{(l)}) = \frac{1}{M} \sum_{j=1}^M \left[k(\mathbf{x}_k^{(j)}, \mathbf{x}_k^{(l)}) \nabla_{\mathbf{x}_k^{(j)}} \log p(\mathbf{x}_k^{(j)}|\theta) + \nabla_{\mathbf{x}_k^{(j)}} k(\mathbf{x}_k^{(j)}, \mathbf{x}_k^{(l)}) \right].
    \end{split}
\end{equation}
In the mean-field \ac{BNN} model introduced in \Cref{sec:gng}, we can just replace $\theta$ by $\mu,\sigma$ in \Cref{eq:stein_grads2}. The computational complexity of the Stein gradient method is $O(M^2)$, which is significantly cheaper than $O(MN)$ when $M << N$. 

\section{Experiments}\label{sec:experiments}

\begin{figure}[h]
    \centering
    \rotatebox[origin=t]{90}{\textbf{Permuted MNIST}}
    \begin{subfigure}{0.48\textwidth}
      \includegraphics[width=\linewidth,trim={.2cm .2cm .2cm .2cm},clip]{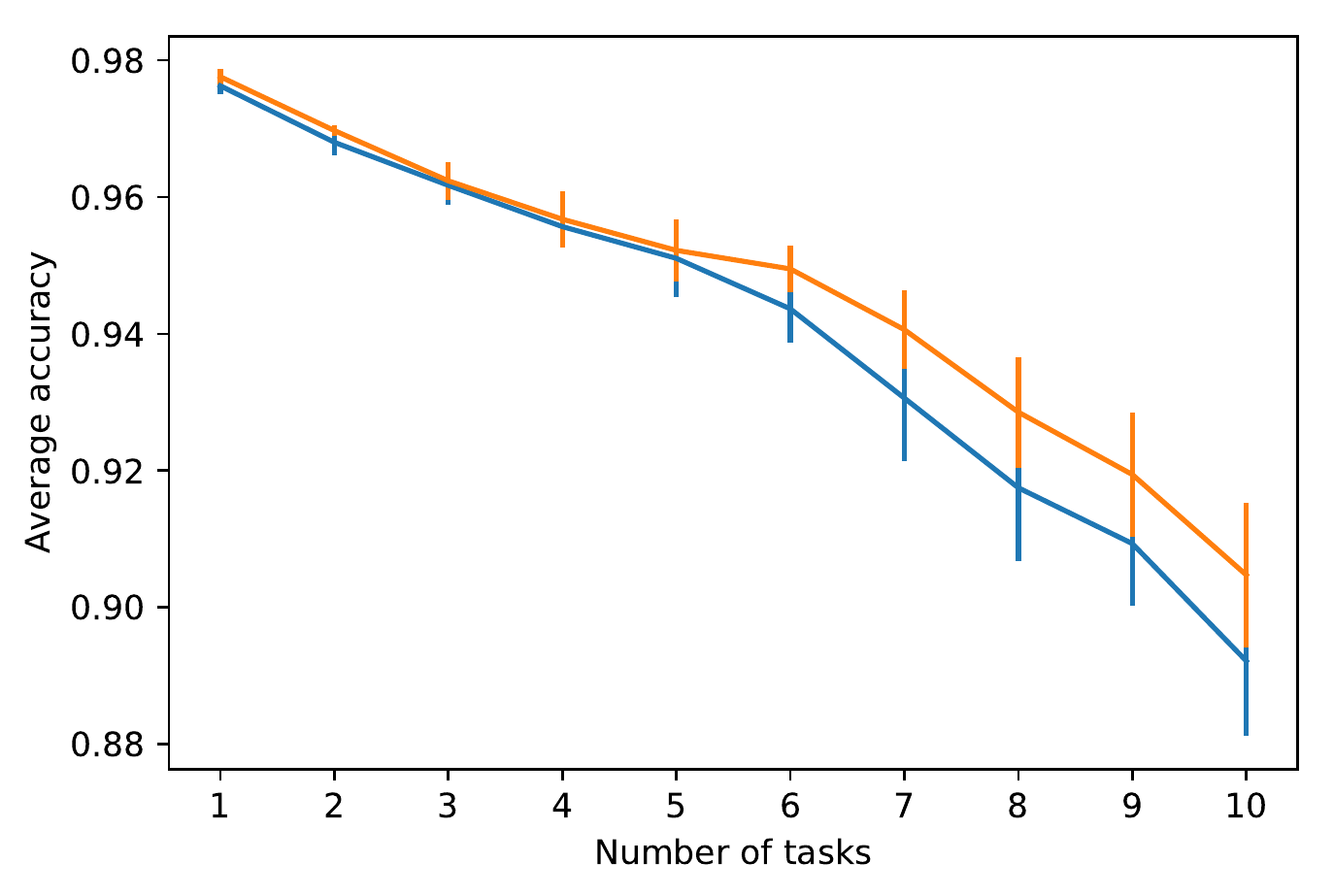}
    \end{subfigure}\hfil
    \begin{subfigure}{0.48\textwidth}
      \includegraphics[width=\linewidth,trim={.65cm .2cm .2cm .2cm},clip]{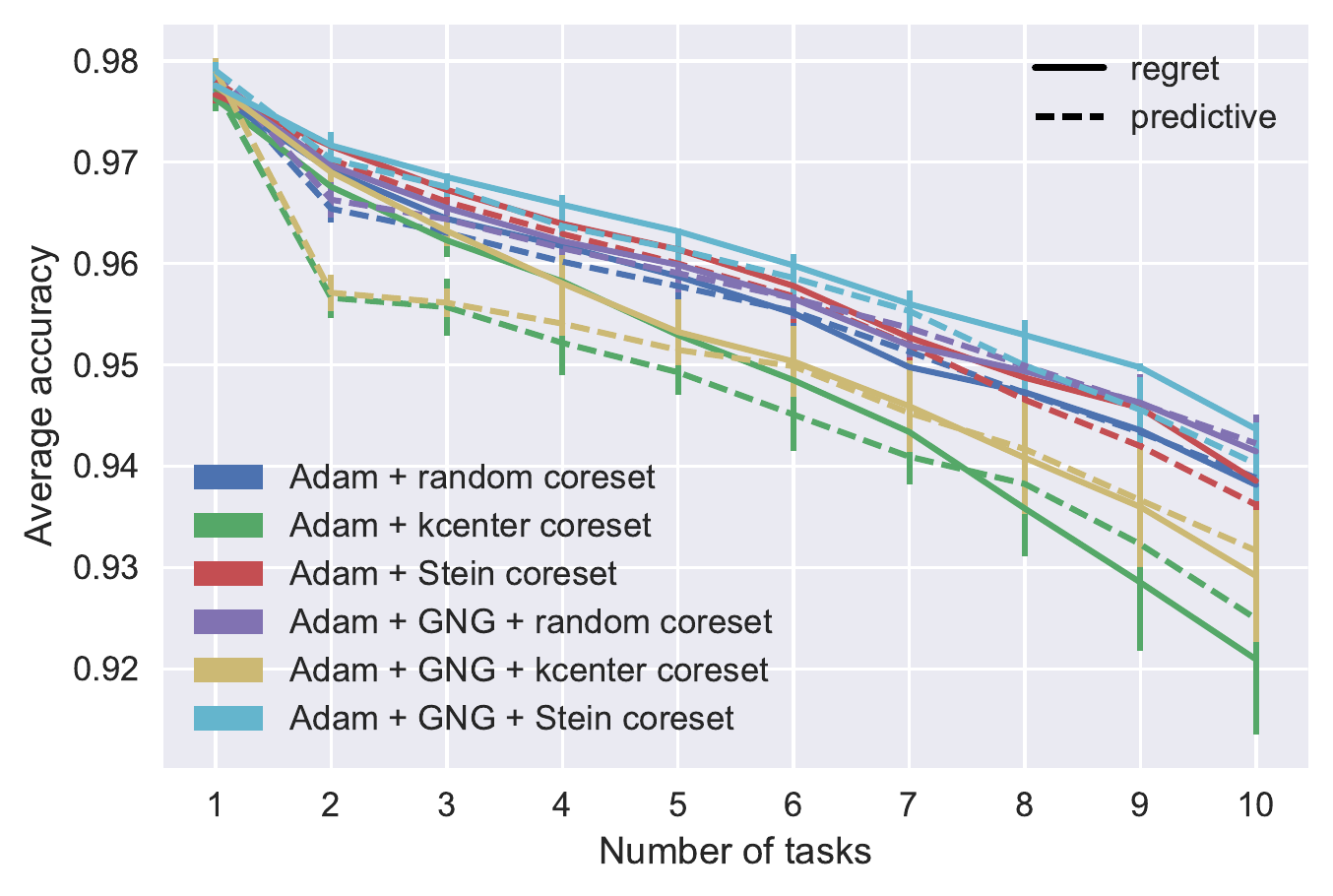}
    \end{subfigure}\hfil

    \medskip

    \rotatebox[origin=t]{90}{\textbf{Split MNIST}}
    \begin{subfigure}{0.48\textwidth}
      \includegraphics[width=\linewidth,trim={.2cm .2cm .2cm .2cm},clip]{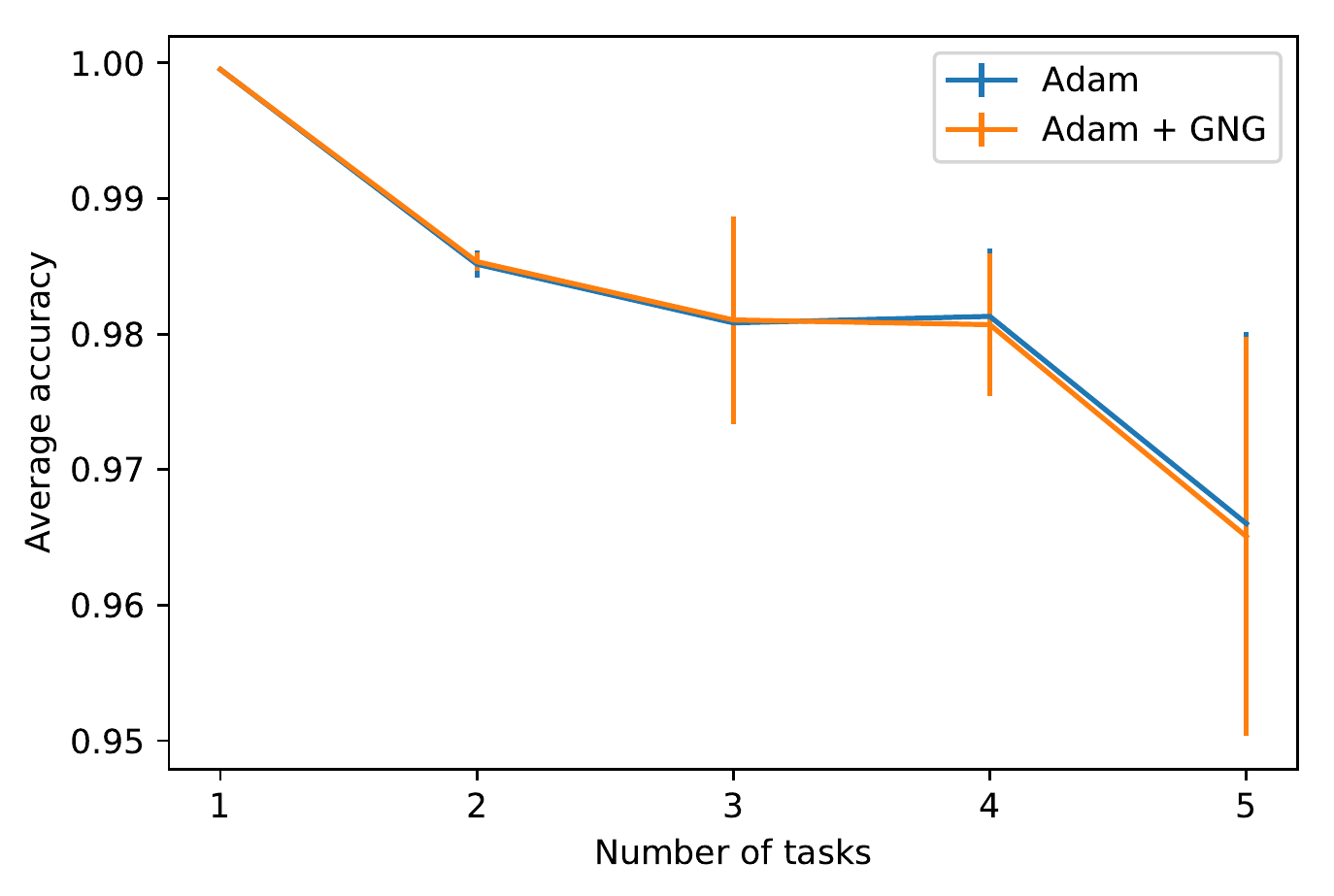}
    \end{subfigure}\hfil
    \begin{subfigure}{0.48\textwidth}
      \includegraphics[width=\linewidth,trim={.65cm .2cm .2cm .2cm},clip]{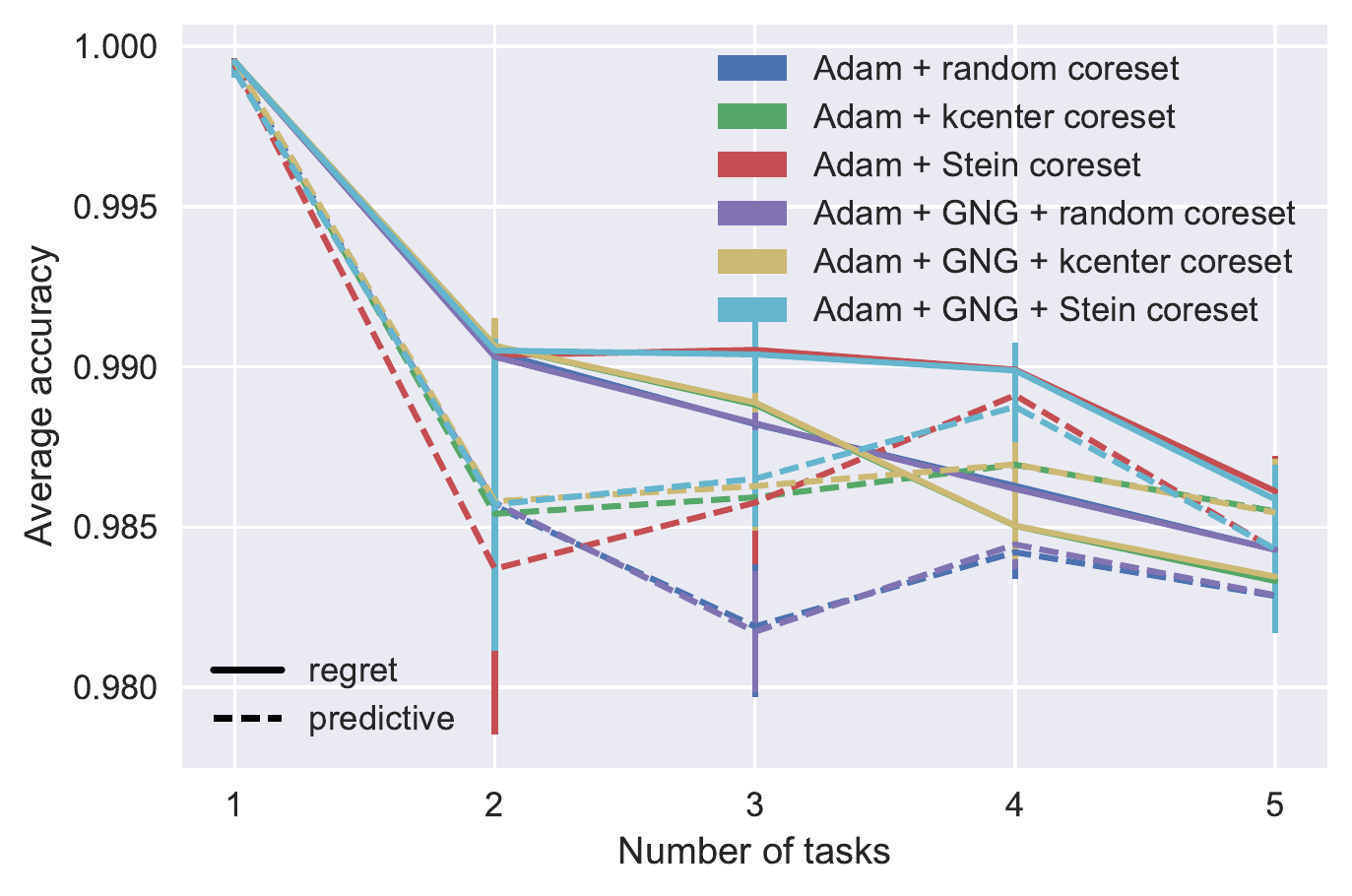}
    \end{subfigure}\hfil
    
    \medskip
    
    \raisebox{0.1cm}{\rotatebox[origin=t]{90}{\textbf{Split Fashion MNIST}}}
    \begin{subfigure}{0.48\textwidth}
      \includegraphics[width=\linewidth,trim={.2cm .2cm .2cm .2cm},clip]{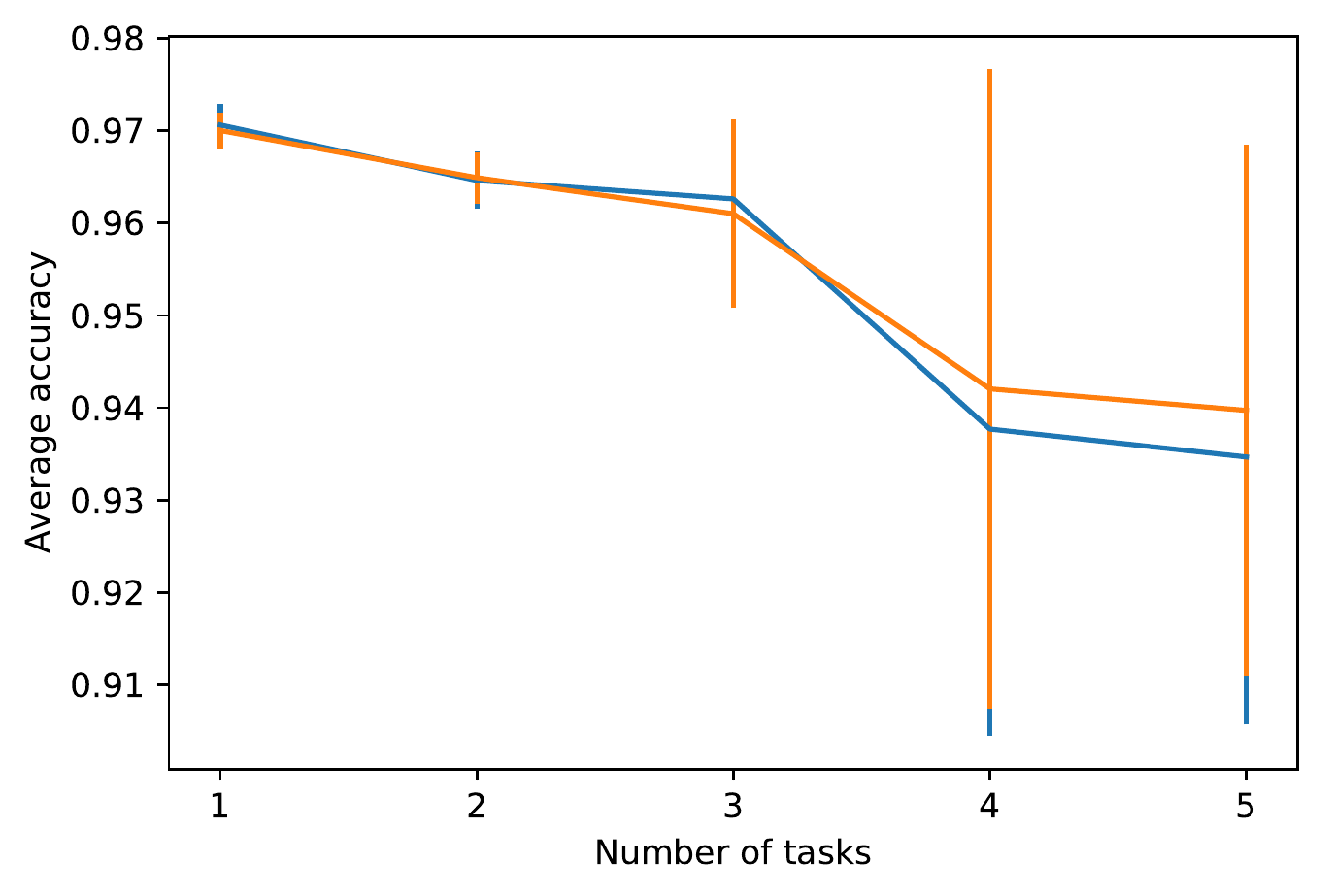}
    \end{subfigure}\hfil
    \begin{subfigure}{0.48\textwidth}
      \includegraphics[width=\linewidth,trim={.65cm .2cm .2cm .2cm},clip]{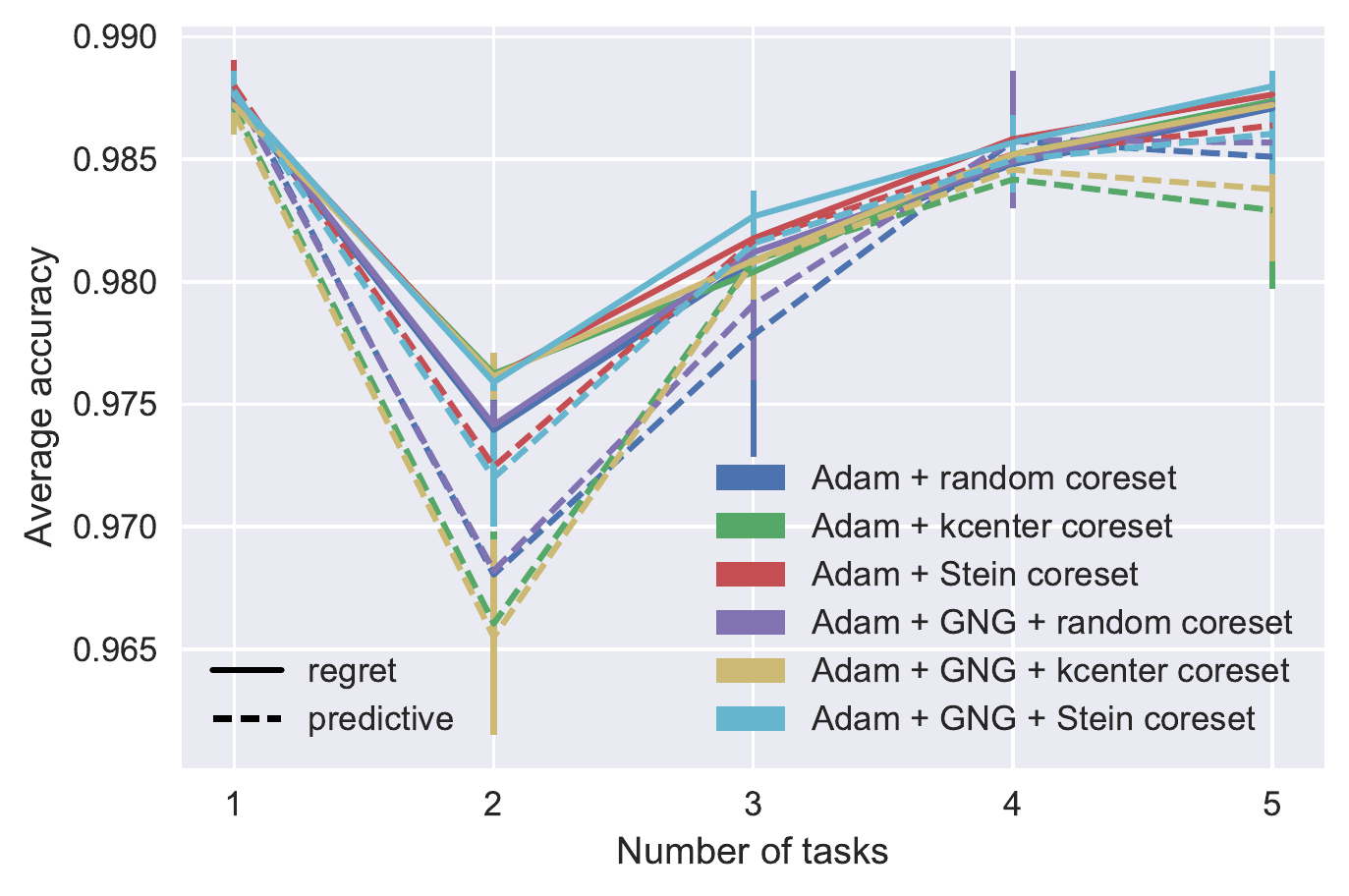}
    \end{subfigure}
\caption{Average accuracy on permuted and split tasks without (left) and with (right) coresets}
\label{fig:gng_coresets}
\end{figure}

We tested \ac{GNG} with Adam in the framework of \ac{VCL} on permuted MNIST \cite{kirkpatrick2017overcoming}, split MNIST \cite{zenke2017continual}, and split fashion MNIST \cite{xiao2017fashion} tasks. We applied a \ac{BNN} with two hidden layers, each layer with 100 hidden units, all split tasks tested using multi-head models \cite{zenke2017continual}. The results are displayed in 
\Cref{fig:gng_coresets} (left column) and the error bars are from 5 runs by different random seeds. In the permuted MNIST task, \ac{GNG} with Adam outperforms standalone Adam. There is no significant difference in split tasks. More details and further analysis can be found in the supplementary material. 

For experiments with coresets, we tested two different usages of coresets. In the first we use coresets to train a predictive model as introduced in \cite{nguyen2017variational} (\Cref{eq:vcl_cs}); in the second we add a regret loss from the coresets to the objective function, which does not require a separate predictive model:
\begin{equation}\label{eq:regret}
    \begin{split}
        \mathcal { L }_t  =  \mathbb { E } _ { q _ { t } ( \mathbf { \theta } ) } \left[ \log p \left( \mathcal{D} _ { t } | \mathbf { \theta } \right) \right] + \mathbb { E } _ { q _ { t } ( \mathbf { \theta } ) } \left[ \log p \left( \mathcal{C} _ { t - 1} | \mathbf { \theta } \right) \right] - \mathrm { KL } \left( q _ { t } ( \mathbf { \theta } ) \| q _ { t - 1 } ( \mathbf { \theta } ) \right),
    \end{split}
\end{equation}
where the second term in \Cref{eq:regret} is regret loss constructed by coresets of  previous tasks $\mathcal{C}_{t-1}$.
We applied a \acs{RBF} kernel in the same manner as described in \cite{liu2016stein} to the Stein gradients and tested the Stein coresets in both permuted and split tasks, comparing with random and $K$-center coresets. The coreset size is $200$ per task in permuted MNIST and $40$ in split tasks, which is the same as used in \cite{nguyen2017variational}. The results are shown in 
\Cref{fig:gng_coresets} (right column). The regret usage of coresets gives better performance in general, and Stein coresets also outperform other two types of coresets in most cases. 

\clearpage
\small
\bibliographystyle{unsrt}
\bibliography{references}

\clearpage

\appendix
\appendixpage
\addappheadtotoc

\section{Comparing different gradient descent algorithms}\label{sec:non_bayes}

\begin{figure}[!htb]
    \centering
    \begin{subfigure}{0.45\linewidth}        \includegraphics[width=\linewidth,trim={.2cm .2cm .5cm .2cm},clip]{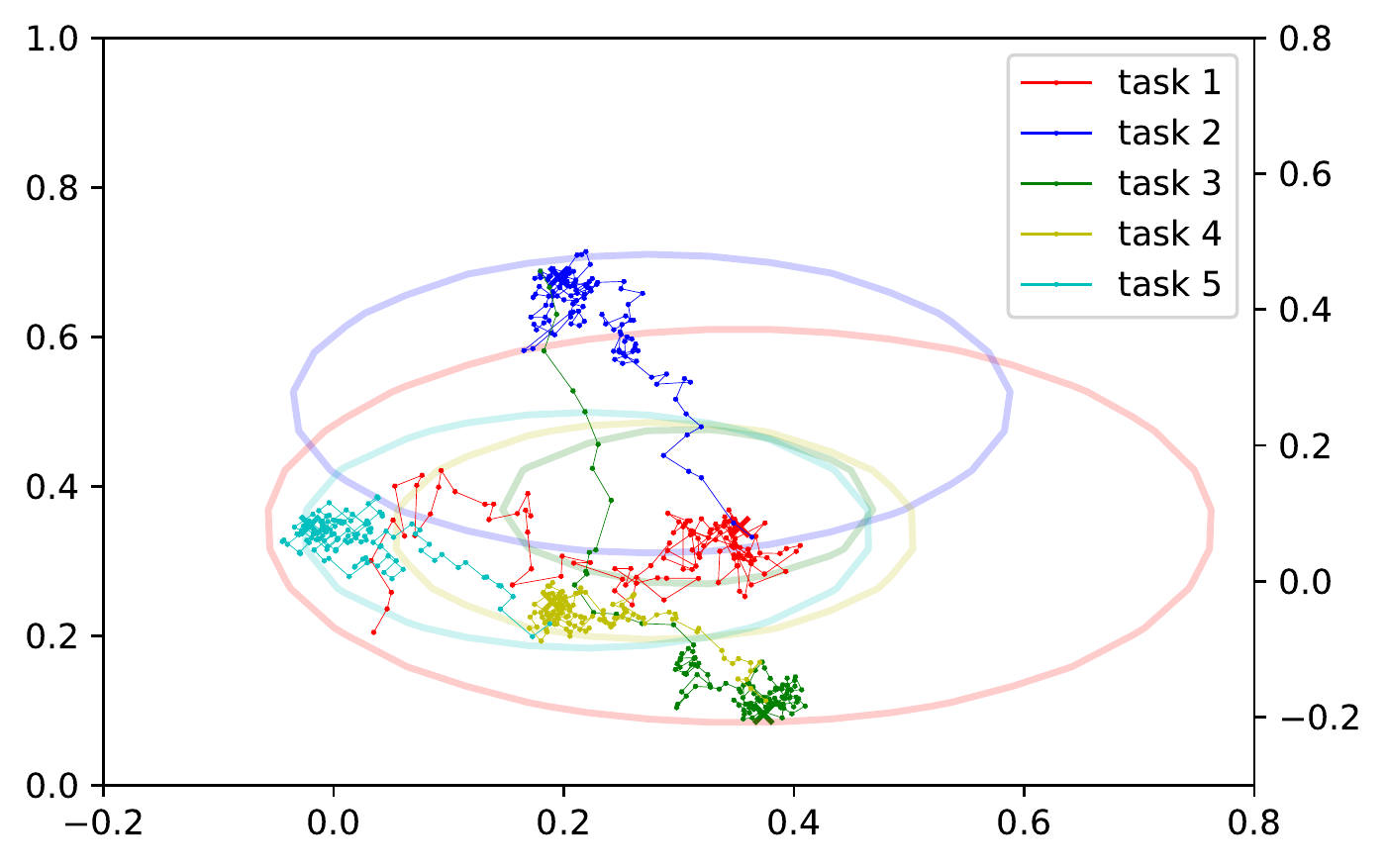}
    \caption{Vanilla SGD}
    \label{fig:sgd_e1}
    \end{subfigure}
     \begin{subfigure}{0.45\linewidth}        \includegraphics[width=\linewidth,trim={.2cm .2cm .5cm .2cm},clip]{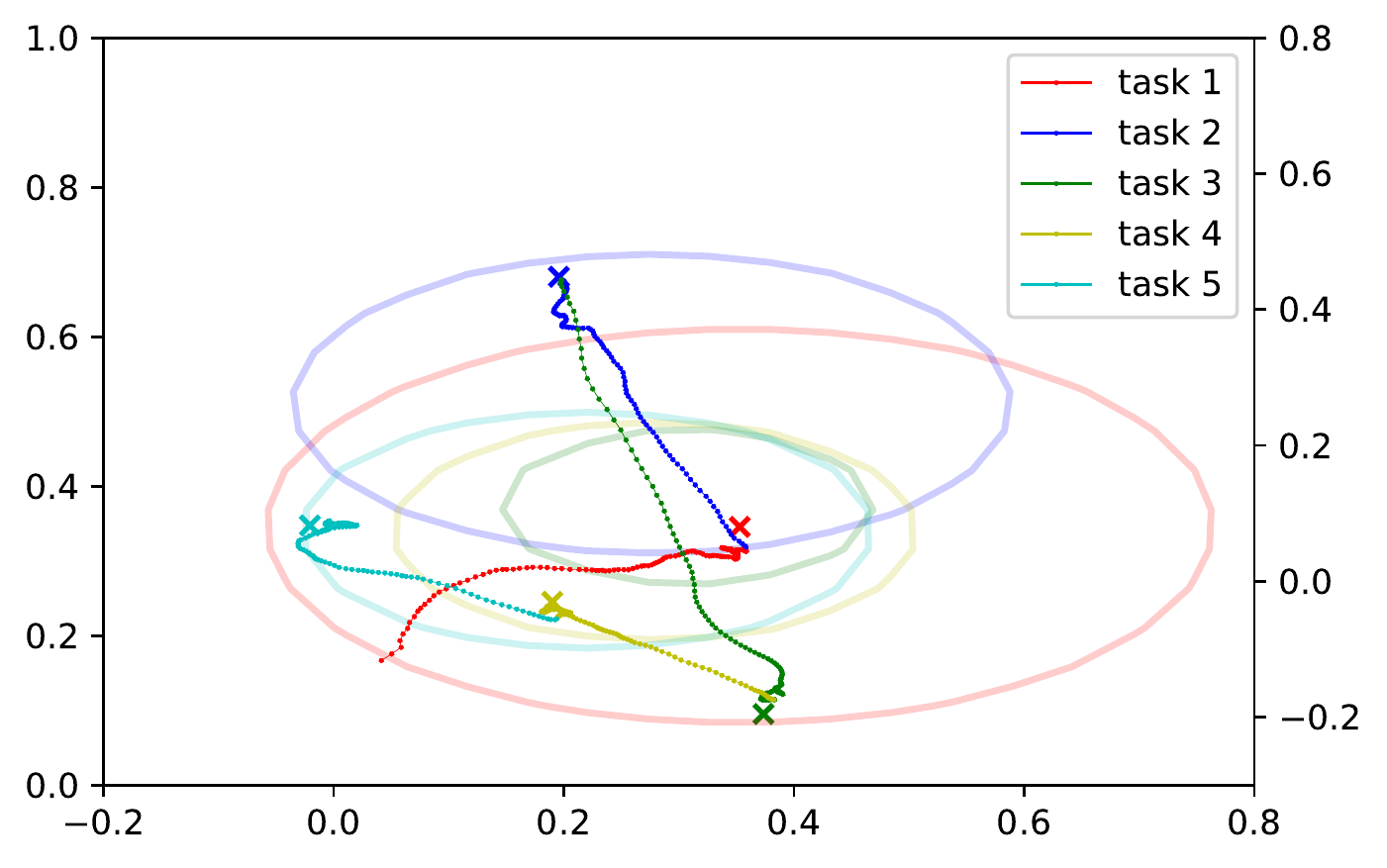}
    \caption{Adam}
    \label{fig:adam_e1}
    \end{subfigure}
    \\%
    \begin{subfigure}{0.45\linewidth}        \includegraphics[width=\linewidth,trim={.2cm .2cm .5cm .2cm},clip]{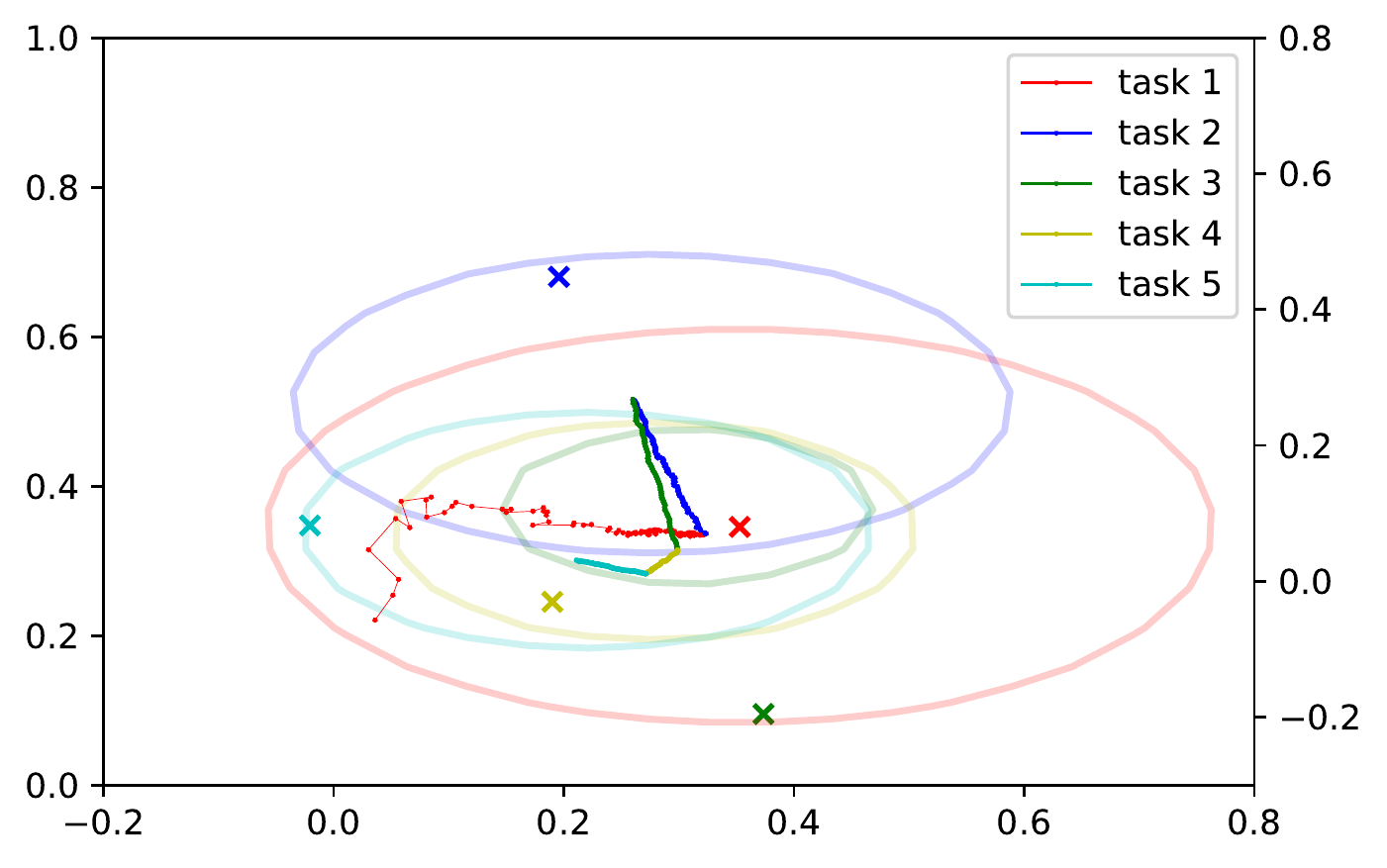}
    \caption{SGD+GNG} 
    \label{fig:sgd_gng_e1}
    \end{subfigure}
     \begin{subfigure}{0.45\linewidth}        \includegraphics[width=\linewidth,trim={.2cm .2cm .5cm .2cm},clip]{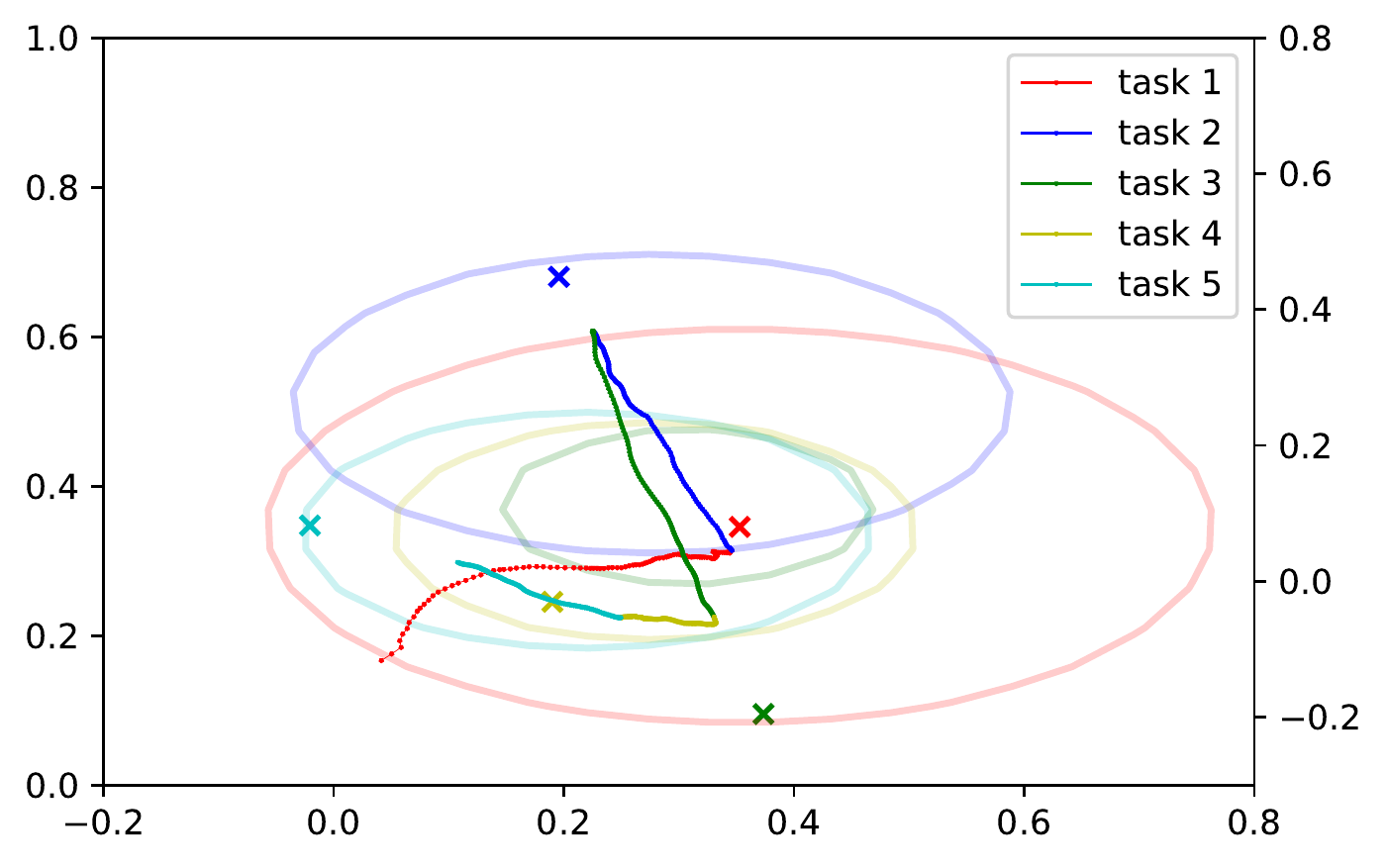}
    \caption{Adam+GNG} 
    \label{fig:adam_gng_e1}
    \end{subfigure}
\caption{Updating trajectory of parameters of 1-dimensional Bayesian linear regression in continual learning. The model is defined as 
$y \sim \mathcal{N}(wx+b,0.1), w \sim \mathcal{N}(\mu_w,\sigma_w^2), b \sim \mathcal{N}(\mu_b,\sigma_b^2)$. The $x$-axis is $\mu_w$ and $y$-axis is $\mu_b$. The contour depicts the average \ac{MSE} over seen tasks, the cross-mark indicates the position of true parameters of each task, different colours represent different tasks. The learning rate is set to $0.001$ for vanilla \ac{SGD} and $0.01$ for all other methods. The initialization of $\sigma_w$ and $\sigma_b$ is set to 
$\sigma_0 = e^{-1}$.}
\label{fig:gd_cmp_e1}
\end{figure}

\begin{figure}[!htb]
    \centering
    \begin{subfigure}{0.45\linewidth}        \includegraphics[width=\linewidth,trim={.2cm .2cm .5cm .2cm},clip]{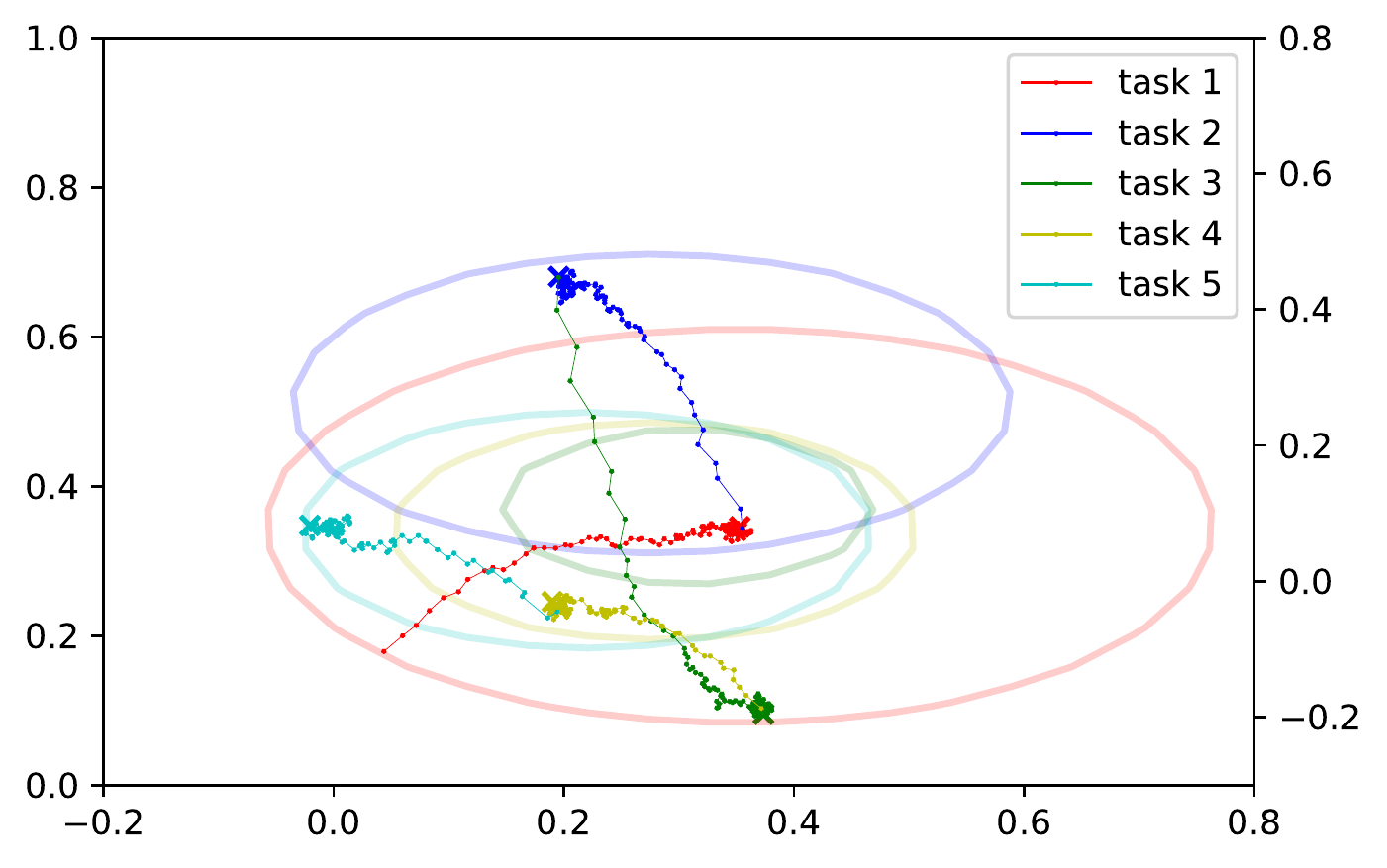}
    \caption{Vanilla SGD}
    \label{fig:sgd_e3}
    \end{subfigure}
     \begin{subfigure}{0.45\linewidth}        \includegraphics[width=\linewidth,trim={.2cm .2cm .5cm .2cm},clip]{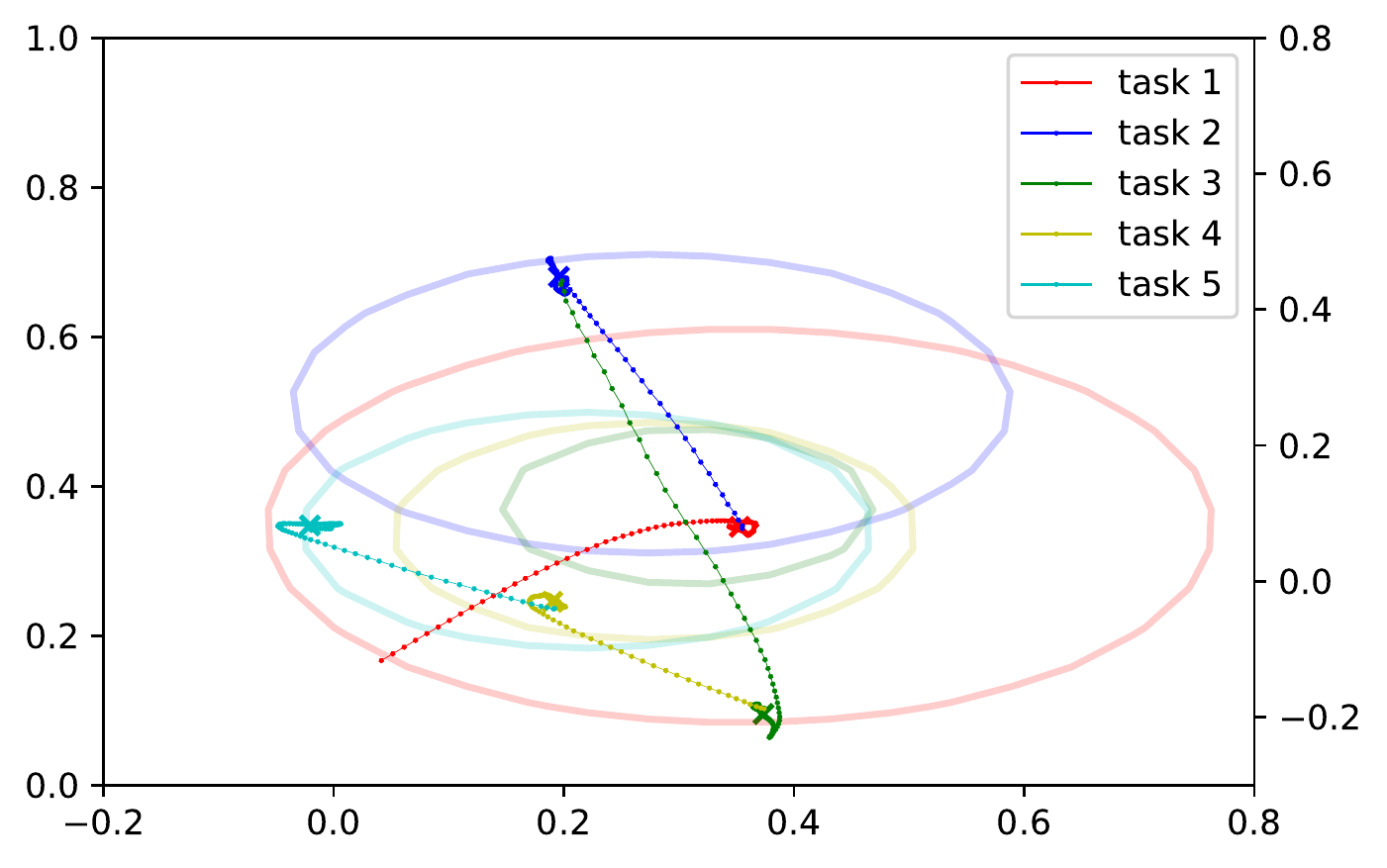}
    \caption{Adam }
    \label{fig:adam_e3}
    \end{subfigure}
    \\%
    \begin{subfigure}{0.45\linewidth}        \includegraphics[width=\linewidth,trim={.2cm .2cm .5cm .2cm},clip]{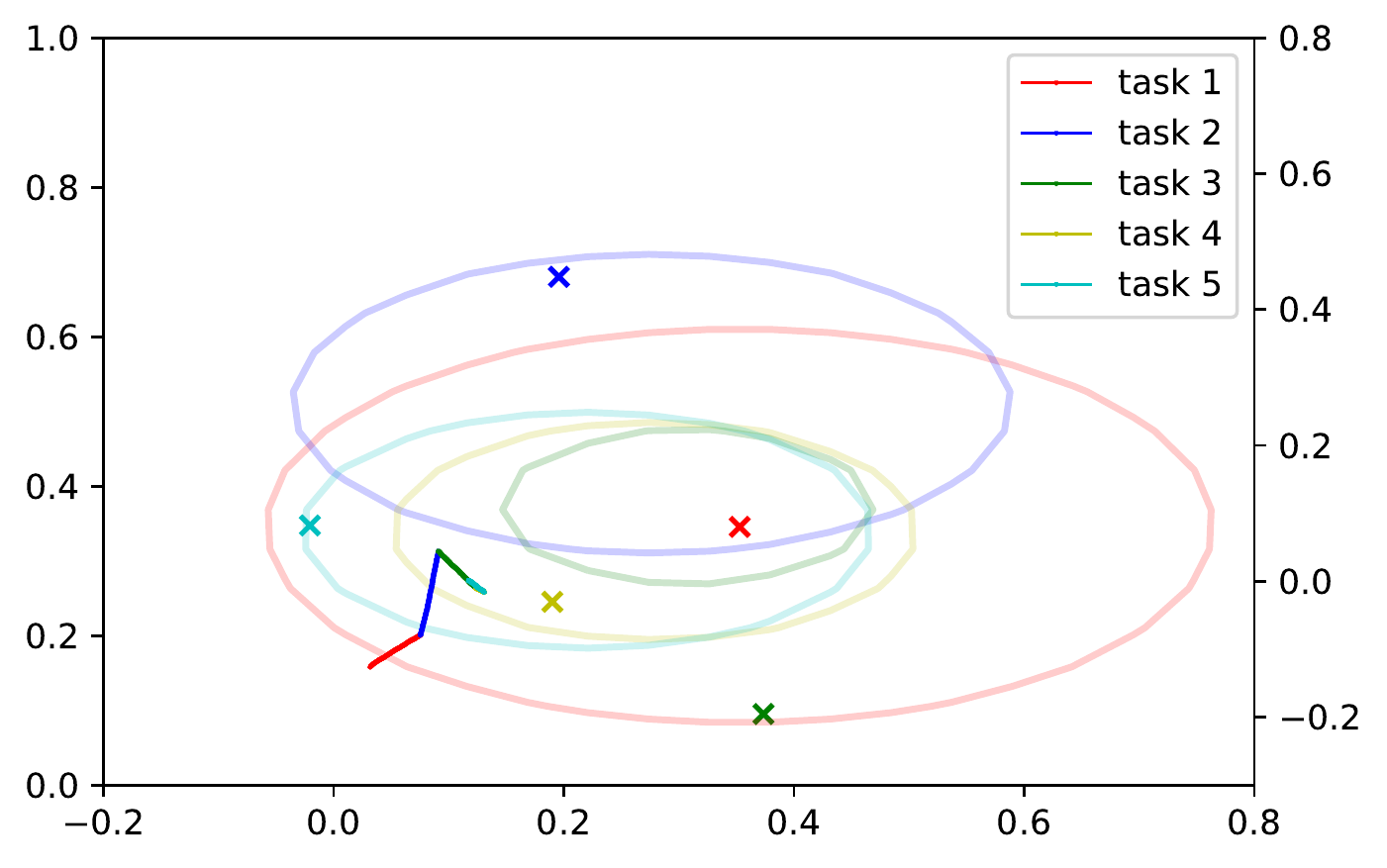}
    \caption{SGD+GNG }
    \label{fig:sgd_gng_e3}
    \end{subfigure}
     \begin{subfigure}{0.45\linewidth}        \includegraphics[width=\linewidth,trim={.2cm .2cm .5cm .2cm},clip]{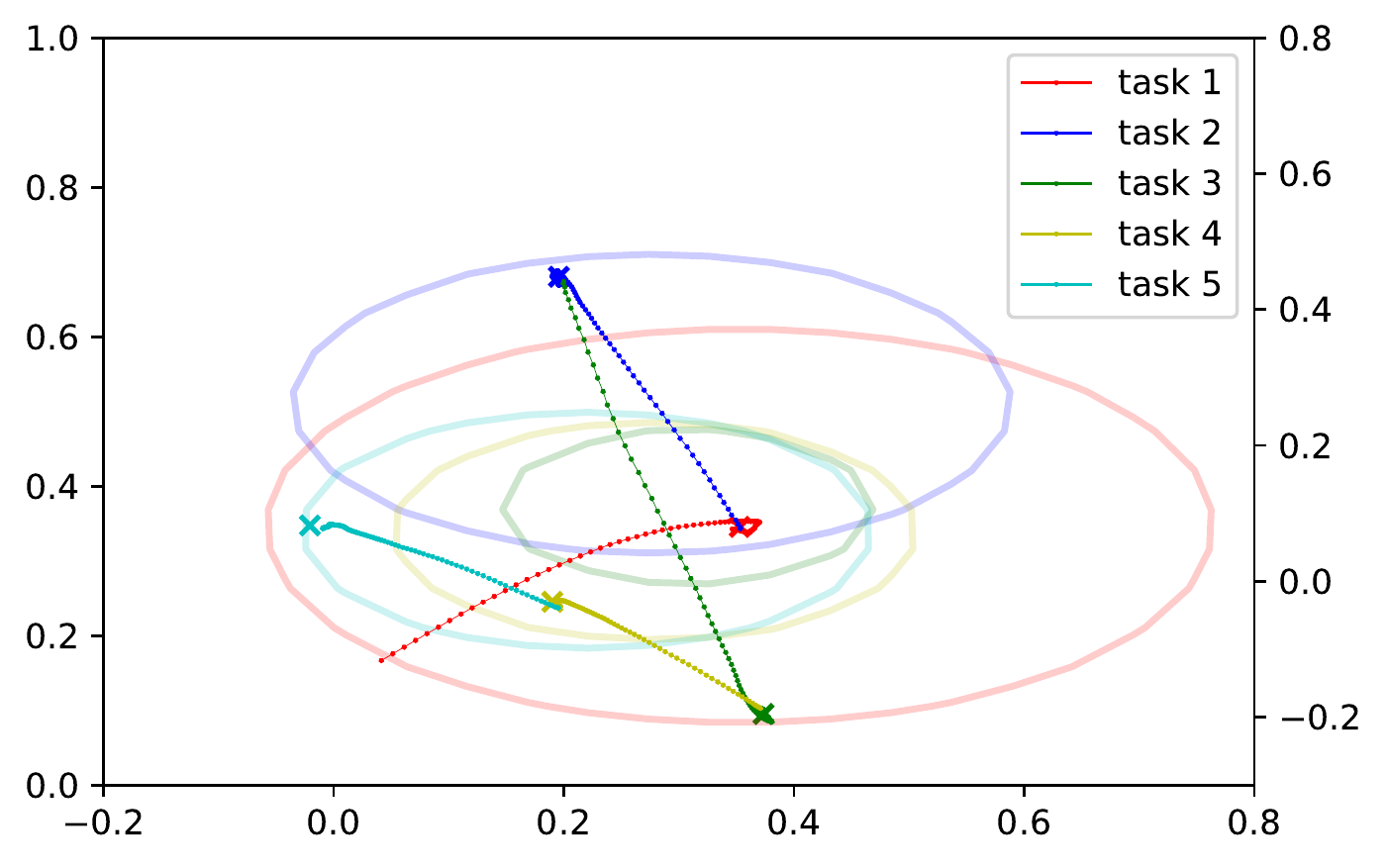}
    \caption{Adam+GNG }
    \label{fig:adam_gng_e3}
    \end{subfigure}
\caption{Parameter trajectory of 1-dimensional Bayesian linear regression in continual learning.
All configurations are the same as in \Cref{fig:gd_cmp_e1} except $\sigma_0 = e^{-3}$.}
\label{fig:gd_cmp_e3}
\end{figure}

\Cref{fig:gd_cmp_e1,fig:gd_cmp_e3} demonstrate how the optimization methods and scale of variance affect parameter updates in an 1-dimensional Bayesian linear regression model of continual learning. In \Cref{fig:adam_gng_e1} the updating steps are smaller than in \Cref{fig:adam_gng_e3}, even when the scale of variance is larger, which is because larger value of initialization $\sigma_0$
results in a larger variance of gradients (see the difference between \Cref{fig:sgd_e1} and \Cref{fig:sgd_e3}, \Cref{fig:adam_e1} and \Cref{fig:adam_e3}), and consequently larger $\text{var}(\sigma_{i,t}^2)$ as well, meaning that the step size of \ac{GNG} decreases according to Equation 7 and 8 in the main content.
In general, \ac{GNG} shows lower variance in parameter updates, and it works better with Adam than with \ac{SGD}.

\section{Further analysis of \aclp{GNG} and Adam experiments}

As one model has a limited capacity, and each different task contains some different information, the ideal case for continual learning is that each new task shares as much information as possible with previous tasks, and occupying as little extra capacity within the \acl{NN} as possible. This is analogous to model compression \cite{louizos2017bayesian}, but one key difference is we want more \emph{free} parameters instead of parameters that are set to zero. For example, there are $k$ independent parameters in a model and the log-likelihood of current task is factorized as:
\begin{equation}
    \begin{split}
        \log p(D_t|\theta_1,\theta_2,\dots,\theta_k) = \sum_{i=1}^k \log p(D_t|\theta_i).
    \end{split}
\end{equation}
If $\theta_1$ is absolutely free for this task, it indicates the following conditional probability is a constant w.r.t. $\theta_1$:
\begin{equation}
    \begin{split}
        \log p(D_t,\theta_2,\dots,\theta_k|\theta_1) 
        &= \log p(D_t|\theta_1) + \sum_{j=2}^k\log p(D_t,\theta_j) 
        = const, \quad \forall \theta_1.
    \end{split}
\end{equation}
This would require
\begin{equation}
    \begin{split}
        \nabla_{\theta_1}\log p(D_t|\theta_1) = 0, \quad \forall \theta_1.
    \end{split}
\end{equation}
Therefore, $\theta_1$ is free to move, then no matter what value of $\theta_1$ is set to in future tasks, it will not affect the loss of previously learned tasks. In realistic situations, $\theta_1$ is very unlikely to be absolutely free. However, it is feasible to maximize the entropy of $\theta_1$, larger entropy indicating more freedom of $\theta_1$. For instance, minimizing KL divergence includes maximizing the entropy of parameters:
\begin{equation}
    \begin{split}
       KL(q_t(\theta)||q_{t-1}(\theta)) = \mathbb{E}_{q_t}[q_{t-1}(\theta)] - H_{q_t}(\theta).
    \end{split}
\end{equation}
On the other hand, it is undesirable to change parameters with lower entropy instead of those with higher entropy while learning a new task, since it could cause a larger loss on previous tasks. 

\begin{figure}[!tb]
    \centering
     \begin{subfigure}{0.495\linewidth} 
    \includegraphics[width=\linewidth,trim={7cm 1cm 6.cm 2.cm},clip]{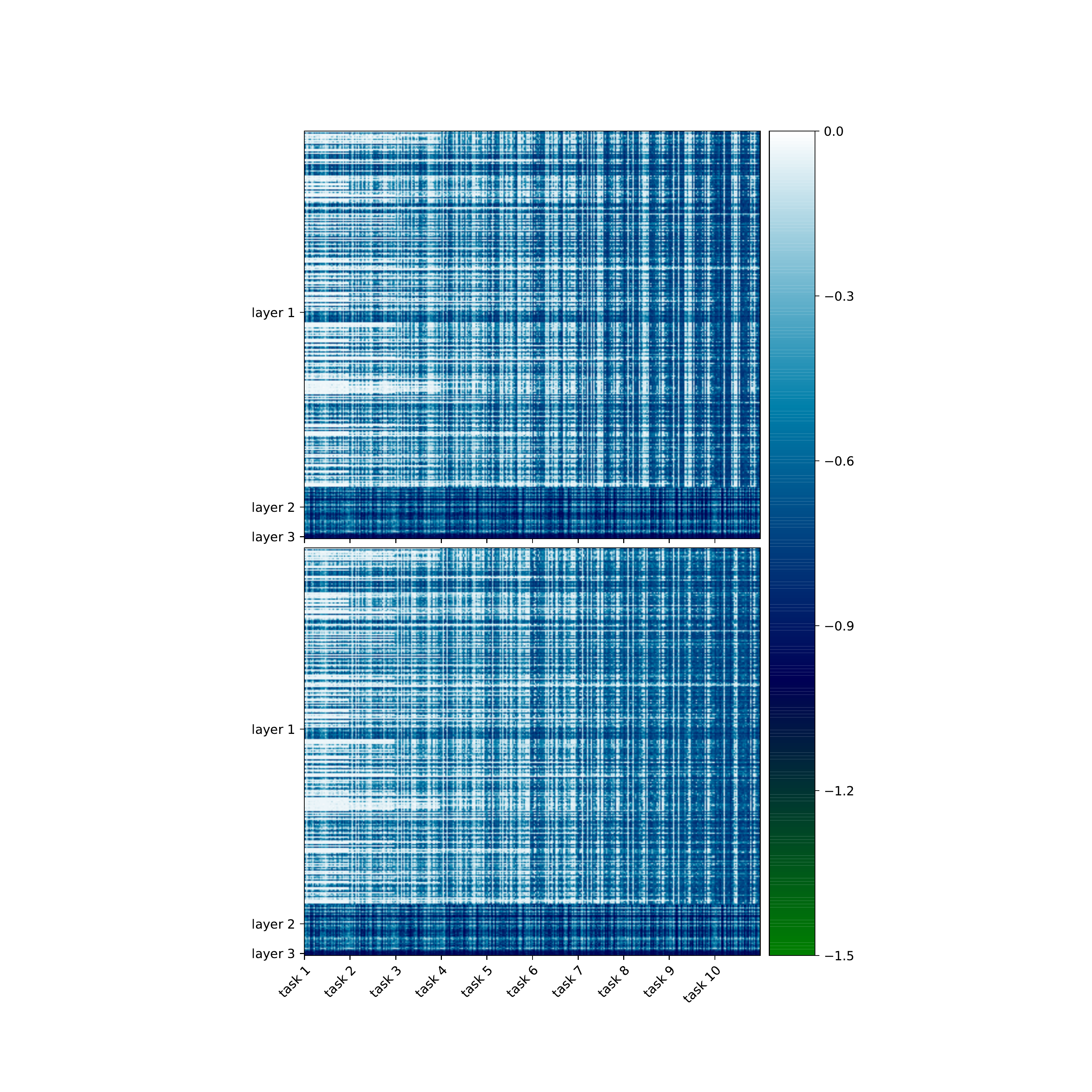}
    \caption{permuted MNIST}
    \label{fig:vheat_permut}
    \end{subfigure}
    \begin{subfigure}{0.495\linewidth} 
    \includegraphics[width=\linewidth,trim={7.cm 1cm 6cm 2.cm},clip]{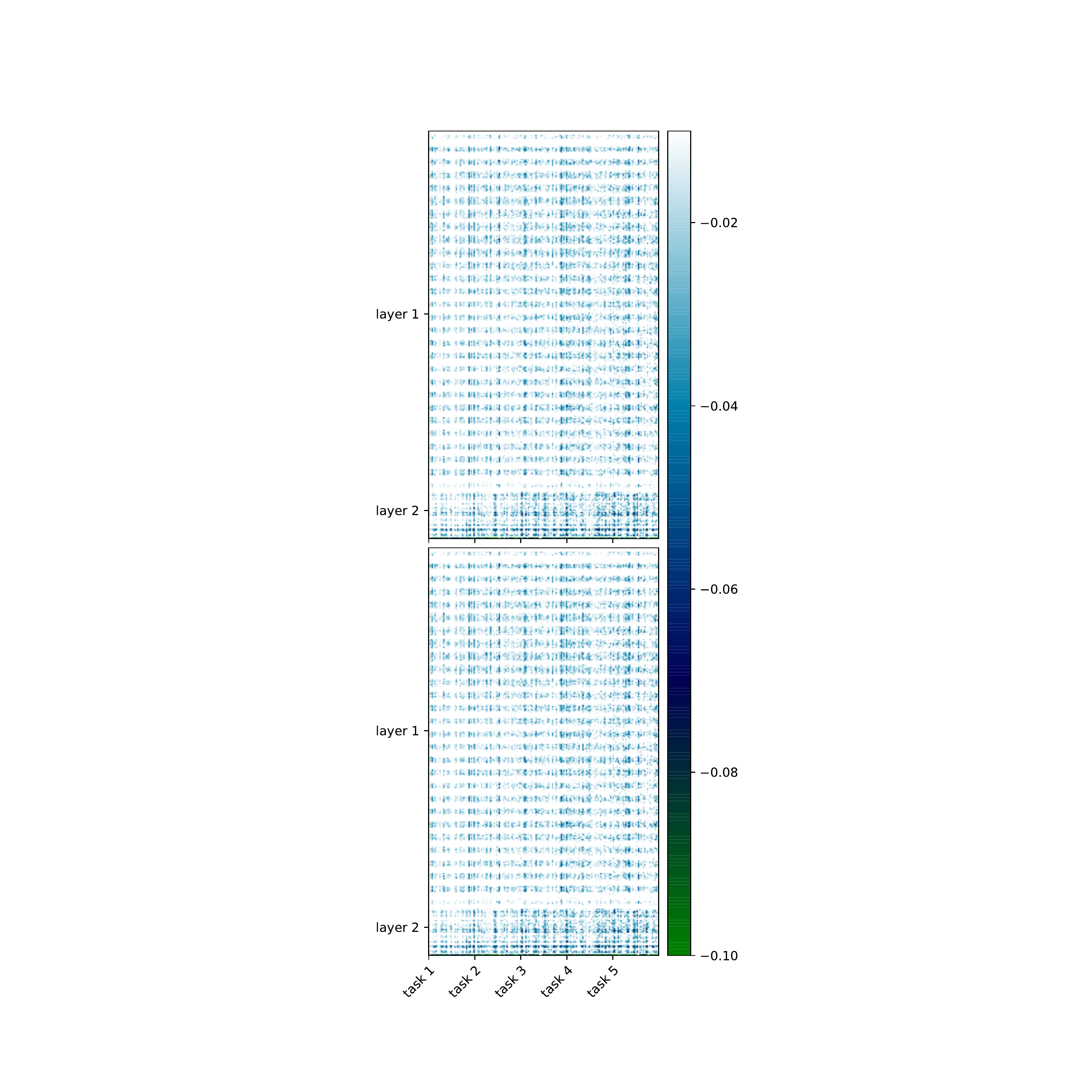}
    \caption{split MNIST}
    \label{fig:vheat_split}
    \end{subfigure}
\caption{Variance changes w.r.t. first task, top row is from models trained by Adam, bottom row is from models trained by Adam + GNG, tested on permuted and split MNIST without coresets. The $x$-axis is concatenated by tasks, the $y$-axis is concatenated by \ac{BNN} layers, as split tasks are tested on multi-head models, so there is no layer 3 in \Cref{fig:vheat_split}.}
\label{fig:vheat}
\end{figure}

The entropy of a Gaussian distribution is defined by its variance alone. In this sense, a larger decrease of variance indicates larger decrease of entropy. To understand why \ac{GNG} works better on permuted MNIST tasks, we visualized how the variances of the weights change in \Cref{fig:vheat} where we normalized all values as below:
\begin{equation}
    \Delta \sigma_{i,t} = \frac{\sigma_{i,t}-\max_{i} \sigma_{i,1}}{\max_{i} \sigma_{i,1}},
\end{equation}
where $\max_{i} \sigma_{i,1}$ is the maximal variance of the first task. 
When the variance of parameters is decreased by learning a new task, the entropy of the model is decreased as well. We can think of it as new information written into the model, so when the model has learned more tasks, the variances of more parameters will have shrunk as shown in \Cref{fig:vheat}. 

In an ideal case, a parameter with larger variance should be chosen to write new information preferentially to avoid erasing information of previous tasks. Therefore, it would be preferred if the dark colour spread more evenly in latter tasks in \Cref{fig:vheat}, and  Adam + \ac{GNG} appears to have this property for the permuted MNIST task (\Cref{fig:vheat_permut}). However, there is no notable difference caused by \ac{GNG} for split MNIST tasks (\Cref{fig:vheat_split}), which consistent with their performance in terms of average accuracy over tasks. The underlying reason needs further investigation.

\end{document}